\documentclass{article}
\usepackage{float}
\usepackage{amsmath,amssymb}
\usepackage{bm}
\usepackage[all]{xy}
\usepackage[colorlinks=true, linkcolor=blue, citecolor=blue, urlcolor=blue]{hyperref}

\title{\bf The Gauss-Markov Adjunction  Provides \\Categorical Semantics of Residuals \\in Supervised Learning}
\author{
Moto Kamiura$^{1,*}$
\\
\\$^1$Institute for Advanced Research and Education, Doshisha University, 
\\1-3 Tatara Miyakodani, Kyotanabe City, Kyoto, 610-0394, Japan
\\
\\$^*$mkamiura@mail.doshisha.ac.jp
\\
}
\date{}

\begin{document}
\maketitle

\begin{abstract}
Enhancing the intelligibility and interpretability of machine learning is a crucial task in responding to the demand for Explicability as an AI principle, 
and in promoting the better social implementation of AI.
The aim of our research is to contribute to this improvement by reformulating machine learning models through the lens of category theory, 
thereby developing a semantic framework for structuring and understanding AI systems.
Our categorical modeling in this paper clarifies and formalizes the structural interplay between residuals and parameters in supervised learning. 
The present paper focuses on the multiple linear regression model, 
which represents the most basic form of supervised learning. 
By defining two Lawvere-enriched categories corresponding to parameters and data, along with an adjoint pair of functors between them, 
we introduce our categorical formulation of supervised learning.
We show that the essential structure of this framework is captured by what we call the Gauss-Markov Adjunction. 
Within this setting, the dual flow of information can be explicitly described as a correspondence between variations in parameters and residuals.
The ordinary least squares estimator for the parameters and the minimum residual are related via the preservation of limits by the right adjoint functor.
Furthermore, we position this formulation as an instance of extended denotational semantics for supervised learning, 
and propose applying a semantic perspective developed in theoretical computer science as a formal foundation for Explicability in AI.
\end{abstract}


\section{Introduction}
Understanding, describing, and explaining the mechanisms of AI at an appropriate level of abstraction has become critically important, 
both for technical research and development and for ethical governance and accountability.
As AI technologies advance and their social implementation progresses, efforts have been made to establish AI principles, 
such as the Asilomar AI Principles \cite{Asilomar2017} and the IEEE Initiative \cite{IEEE2017}. 
These have led to the articulation of five core principles that integrate major ethical frameworks for AI \cite{Floridi+2018, Floridi-Cowls2019}.
Among the five, Explicability stands out as the only principle newly introduced specifically for AI, 
whereas the other four align with the well-known ``four principles of biomedical ethics'' \cite{Beauchamp-Childress1979}.
Explicability integrates both epistemic and ethical dimensions of the transparency required in the context of AI \cite{Floridi+2018}.
Explicability does not merely refer to the disclosure of AI software code \cite{Coeckelbergh2020}, 
but rather requires appropriate attention to the level of abstraction at which explanation is formulated \cite{Floridi2008, Floridi2019}.
From this perspective, it is significant that Ursin et al. \cite{Ursin+2023} positioned Explicability 
as a comprehensive higher-level concept and proposed a four-layered model consisting of disclosure, intelligibility, interpretability, and explainability.
Intelligibility and interpretability, which concern the ability to describe, comprehend, and explain the mechanisms of AI at a suitable level of abstraction, 
play a central role in mediating between the technical details of AI and the ethical demands for transparency. 
These two aspects are therefore essential from both technical and ethical standpoints \cite{Kamiura2025}.

Using category theory to describe AI and machine learning systems is one of the promising approaches to addressing this challenge \cite{Pan2024}. 
Category theory, alongside set theory, provides a foundational framework of abstract algebra that is widely employed across all areas of modern mathematics \cite{MacLane1978}. 
While set theory is based on individual elements, category theory is built upon morphisms. 
These are abstract functions that serve as its primary building blocks. 
A substantial body of research has already explored applying category theory to AI and machine learning.
Many studies, despite having different aims and motivations for using category theory, share a common approach \cite{Abbott+2024, Cruttwell2021, Fong2019, Gavranovic2024, Sheshmani-You2021, Shiebler+2021, Xu+2022}. 
They typically adopt monoidal categories as a foundation and utilize graphical calculi to represent the architectures of learning and inference systems. 
These methods appear to be effective in capturing the modular structures of neural networks and computational graphs, 
as well as in organizing compositional operations and multi-input structures.
This methodological trend may also be shaped by the prior development of monoidal structures and graphical calculi in quantum mechanics, 
where they have been extensively employed to model complex phenomena such as entanglement and quantum teleportation \cite{Heunen-Vicary2019}.
The theoretical resources established in that context are now being reinterpreted and applied to AI research.
Nonetheless, rather than following the mainstream approach based on monoidal categories and graphical calculus, 
we adopt a more conventional categorical framework grounded in standard commutative diagrams.
We make this choice because the effective combination of adjoint functors and standard commutative diagrams 
is likely to yield more satisfactory outcomes in terms of intelligibility and interpretability, 
thereby aligning more closely with the core principles of AI.

In this paper, we focus on the multiple linear regression model, which represents the most fundamental form of supervised learning systems. 
Originally developed as a method in multivariate statistics, 
it now serves as a prototype and foundational template for supervised learning in modern machine learning and AI. 
We formulate regression using category theory by defining two Lawvere-enriched categories corresponding to parameters and data, 
together with an adjoint pair of functors between them. 
We prove that the essential structure of this formulation is captured by what we call the Gauss-Markov Adjunction. 
This framework enables a clear description of the dual flow of information between parameters and residuals, 
expressed as a correspondence between their respective variations. 
The correspondence between the minimum residual and the ordinary least squares (OLS) estimator of the parameters 
is established via the preservation of limits by the right adjoint functor. 
This adjunction-based formulation can also be extended to the case of the minimum norm solution, 
which arises not only in the full row rank setting but also more generally for rank-deficient data matrices requiring the Moore-Penrose inverse.

In the latter part of this paper, we provide an outline of how the Gauss-Markov Adjunction can be connected to semantic modeling. 
In theoretical computer science, denotational semantics is a line of research in which formal meanings of programs 
are clarified by associating symbolic expressions with mathematical models.
In particular, semantic modeling based on category theory is known as categorical semantics \cite{Steingartner2025}.
Although modern AI systems are implemented as software, their internal structure differs fundamentally from that of symbolic or logic-based programs. 
Instead of being composed of discrete syntactic elements, they are constructed as mathematical compositions involving nonlinear functions, calculus, algebra, and statistics.
Therefore, in order to address the semantics of AI systems, we may need to extend the domain of semantic modeling to include machine learning models described in algebraic terms.
The Gauss-Markov Adjunction can be regarded as a representative example of such an extended form of denotational semantics. 
Such an extension of the semantic scope may contribute meaningfully to realizing Explicability as a core principle in AI.

\section{Mathematical and Technical Contexts}
\subsection{Regression}
As preparation for developing a categorical framework for supervised learning,
we summarize key points regarding the conventional regression model.
In this paper, 
uppercase letters are used to denote matrices, and lowercase letters to denote vectors or scalars, unless otherwise specified by the context.
For simplicity, we do not use boldface to denote vectors.

Let $X \in \mathbb{R}^{n \times m}$ be a matrix consisting of $n$ samples of $m$-dimensional explanatory variables (i.e., training data),
$y \in \mathbb{R}^n$ a vector of $n$ samples of a one-dimensional response variable (i.e., target vector),
$r \in \mathbb{R}^n$ a residual vector,
and $a \in \mathbb{R}^m$ a parameter vector.
The pair $(X,y)$ represents a given dataset.
Regression assumes the linear model $y = Xa + r$, and seeks an optimal solution $a=a^*$.
Throughout this paper, we assume the general rank condition ${\rm rank}(X) \le \min(n,m)$, 
thus allowing for rank-deficient data matrices. 
In practical machine learning and statistical analysis, 
appropriate data preprocessing typically makes it reasonable to assume 
${\rm rank}(X) = \min(n,m)$, so that $X$ is row-full-rank or column-full-rank.
For additional information on regression, see {\bf Appendix}.

\subsection{Calibration Parameter}
The multiple regression model can formally be extended by adding an arbitrary parameter vector $b \in \mathbb{R}^n$, resulting in the expression $y = Xa + b + r$.
Note that this $b$ is neither an intercept term, which appears as the first component of the parameter vector $a$, nor a residual term $r$.
Rather, the parameter $b$ represents an explicit calibration term.
Mathematically, $b$ acts as a translation in the affine mapping $\mathcal{F}(a) = Xa +b$.
Since this merely shifts the origin of $y$, and the expression reduces to $z = Xa + r$ by defining $z = y - b$, it may seem that treating $b$ explicitly is unnecessary.

However, in the categorical formulation of regression developed in the following sections,
the inclusion or omission of the calibration term $b$ has a distinct impact on the structure of the corresponding diagrams.
This $b$ appears in the categorical notion of a natural transformation and, in particular, serves to make the unit and counit explicit.
It helps to distinguish functorial operations on objects, morphisms, and functors, thereby revealing a hierarchical semantics that remains implicit in conventional linear-algebraic regression.
If $b$ is disregarded from the outset, the unit and counit become invisible, making them difficult to identify and trace within the structure of the adjunction.
Consequently, retaining $b$ clarifies computations involving functors and facilitates the tracking of adjunction proofs, thereby ensuring categorical consistency.

\subsection{Residual Learning and Structural Duality}
Recent advances in deep learning have underscored the central role of residuals in the training and architecture of complex neural networks. 
The introduction of residual connections in ResNet \cite{He+2016} marked a turning point in deep convolutional network design, 
enabling the training of substantially deeper networks by reformulating the learning task around residual mappings. 
Similarly, the Transformer architecture \cite{Vaswani+2017}, now foundational in modern AI systems, incorporates residual structures as a key mechanism at every layer.
This recurring pattern suggests that residuals are not merely artifacts of statistical estimation but instead constitute a fundamental structural principle in learning systems. 

However, their mathematical and semantic roles remain largely interpreted from an operational or empirical perspective.
In this context, our categorical formulation offers a distinct advantage: 
it formalizes the structural interplay between residuals and parameters as a dual flow of information, captured by the Gauss-Markov Adjunction. 
This perspective not only generalizes classical regression, but also points to a deeper mathematical structure potentially shared across residual-based models in modern machine learning.

While ResNet and Transformer architectures were not designed with categorical semantics in mind, 
the explicit duality we articulate offers a promising direction for understanding 
and perhaps even reengineering these systems based on clearer semantic principles. 
From this perspective, residual learning exemplifies a broader structural motif that our adjunction-based framework seeks to clarify and formally ground.

\subsection{Lawvere-enriched Category}
Two Lawvere-enriched categories corresponding to systems of parameters and data will be defined later in order to describe supervised learning. 
The notion of metric enrichment was introduced by Lawvere \cite{Lawvere1973}, which demonstrated the connection between generalized metric spaces and categories.
A Lawvere-enriched category ${\bf C}$ consists of the objects $c_1,c_2, \cdots \in {\rm Ob}({\bf C})$ together with Hom-objects ${\bf C}(c_1,c_2)={\rm Hom}_{\bf C}(c_1,c_2)$.
These Hom-objects are to be regarded as generalized substitutes for arrows of ordinary categories, and they are objects of the base category ${\bf V} = ([0,\infty], \ge, +, 0)$.
Each Hom-object in ${\bf C}$ can be identified with a distance between objects, ${\bf C}(c_1,c_2)=d(c_1,c_2)\in [0,\infty]$. 
In particular, the identity morphism is assigned by $d(c,c)=0$.
The composition of Hom-objects is defined via the ${\bf V}$-morphism ${\bf C}(c_2,c_3) \otimes {\bf C}(c_1,c_2)\to {\bf C}(c_1,c_3)$, 
which corresponds to the triangle inequality $d(c_2,c_3) + d(c_1,c_2) \ge d(c_1,c_3)$.
A ${\bf V}$-functor $T: {\bf C}\to{\bf D}$ assigns to each object $c \in {\rm Ob}({\bf C})$ an object $Tc \in {\rm Ob}({\bf D})$, 
and is determined by the collection of all morphisms $T_{c_1,c_2}: {\bf C}(c_1,c_2)\to{\bf D}(Tc_1,Tc_2)$, which coherently preserve identities and compositions across the Hom-objects.

Kelly \cite{Kelly1982} defines a monoidal category ${\bf V} = ({\bf V}_0, \otimes, I, a, l, r)$ equipped with natural isomorphisms $a_{c_1c_2c_3}:(c_1 \otimes c_2)\otimes c_3\cong c_1 \otimes (c_2\otimes c_3)$,
$l_c:I\otimes c \cong c$, $r_c: c \otimes I \cong c$, together with an identity element $\iota_c:I\to {\bf C}(c,c)$ for each object $c$.
In the case of a Lawvere-enriched category, these correspond to ${\bf V}_0=[0,\infty]$, $\otimes = +$, and $I = 0$.
Under this definition, a ${\bf V}$-natural transformation can be formulated as follows.
For ${\bf V}$-functors $T,S : {\bf C} \to {\bf D}$, a ${\bf V}$-natural transformation $\alpha : T \to S$ 
is a family of components 
\begin{equation}
\alpha_c : I \to {\bf D}(Tc, Sc)
\label{V-nt-01}
\end{equation}
for each object $c$, satisfying the ${\bf V}$-naturality condition, 
which is expressed by the commutativity of the following diagram:
\begin{equation}
\xymatrix{
I\otimes{\bf C}(c_1,c_2) \ar[rr]^{\alpha_{c_2}\otimes T \qquad}	&&{\bf D}(Tc_2,Sc_2) \otimes {\bf D}(Tc_1,Tc_2)\ar[d]^{composition}\\
{\bf C}(c_1,c_2)\ar[u]^{\cong}_{l^{-1}}\ar[d]_{\cong}^{r^{-1}}				&&{\bf D}(Tc_1,Sc_2)				\\
{\bf C}(c_1,c_2)\otimes I \ar[rr]_{S \otimes \alpha_{c_1} \qquad}	&&{\bf D}(Sc_1,Sc_2) \otimes {\bf D}(Tc_1,Sc_1)\ar[u]_{composition} \; .
}
\label{V-nt-02}
\end{equation}

Applying the definition of a ${\bf V}$-natural transformation (\ref{V-nt-01}) to the case of Lawvere-enriched categories, 
we obtain $\alpha_c : 0 \to {\bf D}(Tc, Sc)$.
This implies $0 \ge {\bf D}(Tc, Sc) = d(Tc, Sc) \ge 0$, and therefore ${\bf D}(Tc, Sc) = 0$, which represents a trivial Hom-object.
Note that such degeneracy is a general feature of metric enrichment.

In Section 3.3, we introduce the $\Lambda$-natural transformation, 
defined as a lax ${\bf V}$-natural transformation, in order to resolve the degeneracy and to reveal the hidden categorical structure of regression.

\section{Categories and Functors for Regression}
Fix an arbitrary matrix $X \in \mathbb{R}^{n \times m}$, 
and let $G \in \mathbb{R}^{m \times n}$ denote the Moore-Penrose inverse of $X$. 
Select an arbitrary vector $b \in \mathbb{R}^n$,  
on which we may impose the constraint $b = b_{\perp} \in {\rm ker}(X^{\top})$ when required. 
Under this constraint, it follows that $X^{\top} b_{\perp} = 0$ and $G b_{\perp} = 0$, 
since ${\rm ker}(G) = {\rm ker}(X^{\top})$.
We refer to $b$ as the \emph{calibration parameter}.

\subsection{Parameter Category and Data Space Category}
As a first step in the mathematical construction, we define two Lawvere-enriched categories and examine their basic properties as follows:

\paragraph{Parameter category}  
We define the Lawvere-enriched category ${\bf Prm}$, whose objects are $m$-dimensional real vectors $a,a_1,a_2,\dots \in \mathbb{R}^m$. 
In particular, all vectors in $\mathbb{R}^m$ arising from linear or affine transformations, such as $Gy$, $Gy-a$, and $Gr$, are included as objects of ${\bf Prm}$.
The Hom-objects are given by 
\begin{equation}
{\rm Hom}_{\bf Prm}(a_1,a_2)=\varphi(a_1,a_2):=\|X(a_2-a_1) \| \in [0,\infty],
\label{Prm-01}
\end{equation}
where the distance function $\varphi$ in ${\bf Prm}$ depends on $X$ and represents a transformed metric, rather than the ordinary Euclidean distance $\|a_2-a_1\|$.
The identity and the composition of the Hom-objects are expressed as follows:
\begin{equation}
\varphi(a,a)=\|X(a-a) \|=0
\label{Prm-02}
\end{equation}
and
\begin{eqnarray}
 &&			\varphi(a_1,a_2) + \varphi(a_2,a_3) \ge \varphi(a_1,a_3) \nonumber \\
 && \Leftrightarrow  \|X(a_2-a_1) \| + \|X(a_3-a_2) \| \ge \|X(a_3-a_1) \| .
\label{Prm-03}
\end{eqnarray}

\paragraph{Data space category}
We define another Lawvere-enriched category ${\bf Data}$, whose objects are $n$-dimensional real vectors $y,y_1,y_2,\dots \in \mathbb{R}^n$. 
In particular, all vectors in $\mathbb{R}^n$ arising from linear or affine transformations, such as $Xa+r$, $y-Xa$, and $XGy$, are included as objects of ${\bf Data}$, in analogy with ${\bf Prm}$.
The Hom-objects given by
\begin{equation}
{\rm Hom}_{\bf Data}(y_1,y_2)=\delta(y_1, y_2):=\|XG(y_2-y_1)\|\in [0,\infty].  
\label{Data-01}
\end{equation}
where the distance function $\delta$ in ${\bf Data}$ depends on $XG$.
The definitions of identity and composition of the Hom-objects also follow analogously.

\bigskip

The categories ${\bf Prm}$ and ${\bf Data}$ are defined as Lawvere-enriched categories. 
Their Hom-objects are defined as distances between parameter objects in ${\bf Prm}$ and between data objects in ${\bf Data}$, and are therefore not reducible to trivial isomorphisms.
Unlike ordinary categories, in which all objects might become isomorphic, 
these enriched categories retain nontrivial structure that is essential to the formulation of the Gauss-Markov adjunction.

\subsection{Affine Forward Functor and Gauss-Markov Functor}
In this section, we define two functors: $\mathcal{F} : {\bf Prm} \to {\bf Data}$ and $\mathcal{G} : {\bf Data} \to {\bf Prm}$.
These functors are determined by the fixed matrix $X$ and the calibration parameter $b$.
While $\mathcal{F}$ may be defined for an arbitrary $b$, in cases where it is useful to emphasize a specific choice of $b$,
we denote it with a subscript as $\mathcal{F}_b$.
Accordingly, $\{\mathcal{F}_b \mid b \in \mathbb{R}^n \}$ denotes the family of functors parameterized by $b$.

\paragraph{Affine forward functor}  
We define the functor $\mathcal{F} : {\bf Prm} \to {\bf Data}$ as follows.  
The functor $\mathcal{F}$ is uniquely determined by the matrix $X$ and a given vector $b \in \mathbb{R}^n$,
and maps each object $a$ and Hom-object $\varphi(a_1,a_2)$ as:
\begin{subequations} \label{F-01}
    \begin{align}
        	&\mathcal{F}:a\to \mathcal{F}a = Xa + b \label{F-01a} \\
		&\mathcal{F}:\varphi(a_1, a_2) \to \delta(\mathcal{F}a_1, \mathcal{F}a_2), \label{F-01b}
    \end{align}
\end{subequations}
where (\ref{F-01b}) implies $\varphi(a_1, a_2)= \|X(a_2-a_1)\| \ge \delta(\mathcal{F}a_1, \mathcal{F}a_2) = \delta(Xa_1 + b, Xa_2 + b) = \|XGX(a_2-a_1)\|= \|X(a_2-a_1)\|$,
which follows from the definition of a ${\bf V}$-functor.
We refer to the functor $\mathcal{F}$ defined in (\ref{F-01a}) and (\ref{F-01b}) as the \emph{affine forward functor}. 
This name reflects the fact that $\mathcal{F}$ is an affine map with respect to the objects $a$ of the category ${\bf Prm}$, 
and that it generates the forward model for the input data $X$ based on the parameter $a$.
We can verify that $\mathcal{F}$ preserves both identity and composition of Hom-objects in the category ${\bf Data}$ as follows:
{\bf Identity}:
$\varphi(a, a)=\|X(a-a)\|=0$ and $\delta(\mathcal{F}a, \mathcal{F}a) = \delta(Xa + b, Xa + b) = \|XGX(a-a)\|=0$.
{\bf Composition:}
$\varphi(a_1, a_2)+\varphi(a_2, a_3) = \|X(a_2-a_1)\| + \|X(a_3-a_2)\|\ge \|X(a_3-a_1)\| = \varphi(a_1, a_3)$
and 
$
\delta(\mathcal{F}a_1, \mathcal{F}a_2) + \delta(\mathcal{F}a_2, \mathcal{F}a_3)  
= \delta(Xa_1 + b, Xa_2 + b) + \delta(Xa_2 + b, Xa_3 + b) 
= \|XGX(a_2-a_1)\| + \|XGX(a_3-a_2)\| \ge \|XGX(a_3-a_1)\| = \delta(\mathcal{F}a_1, \mathcal{F}a_3)
$
.

\paragraph{Gauss-Markov functor}
We define the functor $\mathcal{G}: {\bf Data} \to {\bf Prm}$ as follows. 
It is induced by the Moore-Penrose inverse $G$ of $X$.
This functor maps each object $y$ and Hom-object $\delta(y_1,y_2)$  as:
\begin{subequations} \label{G-01}
    \begin{align}
	&\mathcal{G}:y\to \mathcal{G}y=G y  \label{G-01a} \\
	&\mathcal{G}:\delta(y_1, y_2) \to \varphi(\mathcal{G}y_1, \mathcal{G}y_2) ,  \label{G-01b}
    \end{align}
\end{subequations}
where (\ref{G-01b}) implies $\delta(y_1, y_2)=\|XG(y_2-y_1)\| \ge \varphi(\mathcal{G}y_1, \mathcal{G}y_2) = \varphi(Gy_1, Gy_2) = \|XG(y_2-y_1)\|$.
We refer to the functor $\mathcal{G}$ defined in (\ref{G-01a}) and (\ref{G-01b}) as the \emph{Gauss-Markov functor}.
This name is derived from the Gauss-Markov theorem, which proves that the least squares estimator of the parameters in a linear regression model 
is the best linear unbiased estimator, using the matrix $G$ as a key element in the proof.
The Gauss-Markov functor reconstructs the parameter system from observed data.
We can verify that $\mathcal{G}$ preserves both identity and composition of Hom-objects in the category ${\bf Prm}$ as follows:
$\delta(y, y) =\|XG(y-y)\|=0 \ge \varphi(\mathcal{G}y, \mathcal{G}y) = \varphi(Gy, Gy) = \|XG(y-y)\|=0$
and
$
\delta(y_1, y_2)+\delta(y_2, y_3) 
\ge \varphi(\mathcal{G}y_1, \mathcal{G}y_2) + \varphi(\mathcal{G}y_2, \mathcal{G}y_3)  
= \varphi(Gy_1, Gy_2) + \varphi(Gy_2, Gy_3) 
= \|XG(y_2-y_1)\| + \|XG(y_3-y_2)\| \ge \|XG(y_3-y_1)\| = \varphi(\mathcal{G}y_1, \mathcal{G}y_3) 
$
.

\bigskip
From this point onward, we may occasionally omit parentheses and the composition operator for the sake of notational simplicity, 
writing expressions such as $\mathcal{F}a$, $\mathcal{G}y$, and $\mathcal{GF}$ in place of $\mathcal{F}(a)$, $\mathcal{G}(y)$, and $\mathcal{G} \circ \mathcal{F}$.

\subsection{$\Lambda$-Natural Transformation}
As shown in Section 2.4, a ${\bf V}$-natural transformation (\ref{V-nt-01}) generally degenerates to $0$ in the case of Lawvere-enriched categories.
In this section, we introduce the $\Lambda$-natural transformation, 
defined as a lax ${\bf V}$-natural transformation, in order to resolve this degeneracy and to reveal the hidden categorical structure of regression.

First, we define the following specific Hom-object,
\begin{equation}
\Lambda := \|XGb\|.
\label{Lambda-00}
\end{equation}
This equation means that $\Lambda$ is a canonical lifting of the calibration parameter $b$, from a mere parameter vector to a Hom-object.
From the affine forward functors, $\mathcal{F}_0 :a \to Xa$ and $\mathcal{F}_b:a \to Xa+b$, for any $a \in {\rm Ob}({\bf Prm})$,
we obtain the following equation,
\begin{equation}
{\rm Hom}_{\bf Data}(\mathcal{F}_0a, \mathcal{F}_ba) = \|XG\{(Xa+b)-Xa\}\|=\|XGb\|=\Lambda.
\label{Lambda-AF-00}
\end{equation}
Therefore, $\Lambda$ can be interpreted as the internal metric deviation between the linear and affine components of $\mathcal{F}$.
It has a desirable property that is preserved under the mapping by the Gauss-Markov functor $\mathcal{G}$:
\begin{equation}
{\rm Hom}_{\bf Prm}(\mathcal{GF}_0a, \mathcal{GF}_ba) = \|X\{G(Xa+b)-GXa\}\|=\|XGb\|=\Lambda.
\label{Lambda-GM-00}
\end{equation}
Therefore, the following diagram commutes:
\begin{equation}
\xymatrix{
					& \delta(\mathcal{F}_0a, \mathcal{F}_ba) \ar[dd]^{\mathcal{G}} \\
\Lambda \ar[ru]^{=}\ar[rd]_{=}& 											\\
					& \varphi(\mathcal{GF}_0a, \mathcal{GF}_ba) \;.
}
\end{equation}
If $b=b_{\perp}$, then $\Lambda_{b_{\perp}}=\|XGb_{\perp}\|=0=I$, and the above diagram degenerates.

Under this definition, a $\Lambda$-natural transformation can be formulated as follows.
For affine forward functors, $\mathcal{F}_1 :a \to Xa + b_1$ and $\mathcal{F}_2:a \to Xa+b_2$, a $\Lambda$-natural transformation $\lambda : \mathcal{F}_1 \to \mathcal{F}_2$ 
is a family of components 
\begin{equation}
\lambda_a : \Lambda \to \delta(\mathcal{F}_1a, \mathcal{F}_2a) = \|XG(b_2-b_1) \|
\label{Lambda-nt-01}
\end{equation}
for each object $a$, satisfying the \emph{$\Lambda$-naturality} condition, 
which is expressed by the commutativity of the following diagram:
\begin{equation}
\xymatrix{
\Lambda \otimes \varphi(a_1,a_2) \ar@<0.5ex>[dd] \ar[rr]^{\lambda_{a_2}\otimes \mathcal{F}_1 \qquad}	&&\delta(\mathcal{F}_1 a_2, \mathcal{F}_2 a_2) \otimes \delta(\mathcal{F}_1 a_1,\mathcal{F}_1 a_2)\ar[d]^{composition}\\
											&&\delta(\mathcal{F}_1 a_1, \mathcal{F}_2 a_2)				\\
\varphi(a_1,a_2) \otimes \Lambda \ar@<0.5ex>[uu]^{\cong} \ar[rr]_{\mathcal{F}_2 \otimes \lambda_{a_1} \qquad}	&&\delta(\mathcal{F}_2 a_1, \mathcal{F}_2 a_2) \otimes \delta(\mathcal{F}_1 a_1,\mathcal{F}_2 a_1)\ar[u]_{composition} \; .
}
\label{Lambda-nt-02}
\end{equation}
Equations (\ref{Lambda-nt-01}) and (\ref{Lambda-nt-02}) together define the $\Lambda$-natural transformation
as a lax ${\bf V}$-natural transformation, inspired by (\ref{V-nt-01}) and (\ref{V-nt-02}).
If $b \ge b_2 - b_1$, then the $\Lambda$-natural transformation $\lambda_a$ exists,
thus avoiding the degeneracy inherent in the ${\bf V}$-natural transformation of Lawvere-enriched categories.

\section{Gauss-Markov Adjunction}
In this section, we prove the following statement:

\paragraph{Proposition (GM-1)}
For the affine forward functor $\mathcal{F}$ and the Gauss-Markov functor $\mathcal{G}$, there exists a $\Lambda$-natural transformation
\begin{equation}
\Phi_{\Lambda} : {\rm Hom}_{\bf Data}(\mathcal{F}a, y) \to {\rm Hom}_{\bf Prm}(a, \mathcal{G}y).
\label{adj-01a}
\end{equation}
In particular, for $\mathcal{F}_{b_{\perp}}$, constrained by $b = \,^{\forall}b_{\perp} \in {\rm ker}(X^{\top})$, 
there exists a ${\bf V}$-natural isomorphism
\begin{equation}
\Phi_{\rm GM} : {\rm Hom}_{\bf Data}(\mathcal{F}_{b_{\perp}}a, y) \cong {\rm Hom}_{\bf Prm}(a, \mathcal{G}y).
\label{adj-01b}
\end{equation}
This adjunction between $\mathcal{F}_{b_{\perp}}$ and $\mathcal{G}$ is referred to as the \emph{Gauss-Markov Adjunction}. \hfill $\square$

\bigskip
To prove the proposition {\bf (GM-1)}, it is necessary and sufficient, according to Awodey \cite{Awodey2010}, 
to establish the following {\bf (GM-2)} and {\bf (GM-3)}.
In the present context, these reformulations of the proposition serve to clarify the unit $\omega_a$ and counit $\varepsilon_y$ associated with this adjunction.

\paragraph{Proposition (GM-2)}
There exists a $\Lambda$-natural transformation $\bar \omega: \mathcal{I} \to \mathcal{GF}$,
\begin{equation}
\bar \omega_a : \Lambda \to {\rm Hom}_{\bf Prm}(a,\mathcal{G}\mathcal{F}a) = \varphi(a,\mathcal{GF}a)= \omega_a,
\label{adj-02-unit}
\end{equation}
which we refer to as a \emph{$\Lambda$-unit}, with the following property.
For any objects $a \in {\rm Ob}({\bf Prm})$ and $y \in {\rm Ob}({\bf Data})$, and for a Hom-object $\varphi(a,\mathcal{G}y)$, 
there exists a Hom-object $\delta(\mathcal{F}a,y)$ such that the following diagram commutes:
\begin{equation}
\xymatrix{
\Lambda \otimes \delta(\mathcal{F} a, y) \ar@<0.5ex>[dd] \ar[rr]^{\bar \omega_a \otimes \mathcal{G} \qquad}	&& \varphi(a, \mathcal{GF} a) \otimes \varphi(\mathcal{GF} a, \mathcal{G} y )\ar[d]^{composition}\\
																		&&\varphi(a, \mathcal{G} y)				\\
\delta(\mathcal{F}a, y) \otimes \Lambda \ar@<0.5ex>[uu]^{\cong} \ar[rr]_{\mathcal{G} \otimes \bar \omega_a \qquad}	&& \varphi(\mathcal{GF} a, \mathcal{G} y ) \otimes \varphi(a, \mathcal{GF} a) \ar[u]_{composition} \; ,
}
\label{adj-02a2}
\end{equation}
where the right vertical arrow implies
\begin{equation}
\varphi(\mathcal{G}\mathcal{F}a,\mathcal{G}y) + \omega_a \;\ge\; \varphi(a,\mathcal{G}y).
\label{adj-02a1}
\end{equation}

In particular, if we select $b = \,^{\forall}b_{\perp} \in {\rm ker}(X^{\top})$, 
then $\omega_{a,b_{\perp}}=0$, $\Lambda_{b_{\perp}}=I=0$, 
the equality in (\ref{adj-02a1}) holds,
and $\delta(\mathcal{F}a,y)$ is uniquely determined for any $a$, 
$y$, and $\varphi(a,\mathcal{G}y)$; that is,
\begin{equation}
\varphi(\mathcal{G}\mathcal{F}_{b_{\perp}}a,\mathcal{G}y) + \omega_{a,b_{\perp}}
\;=\;
\varphi(a,\mathcal{G}y),
\label{adj-02b1}
\end{equation}
which implies the isomorphism,
\begin{equation}
{\rm Hom}_{\bf Prm}(\mathcal{G}\mathcal{F}_{b_{\perp}}a,\mathcal{G}y)\,\otimes\,{\rm Hom}_{\bf Prm}(a,\mathcal{G}\mathcal{F}_{b_{\perp}}a)
\;\xrightarrow{\;\cong\;}\;
{\rm Hom}_{\bf Prm}(a,\mathcal{G}y).
\label{adj-02b2}
\end{equation}
This is the universal mapping property of the unit $\omega_{a,b_{\perp}}$, in the sense of ${\bf V}$-natural transformation.
\hfill $\square$

\paragraph{Proof (GM-2)}
From the definitions of the functors and the Hom-objects, we obtain
$\varphi(a, \mathcal{G}y)=\|X(Gy-a)\|$,
$\mathcal{G}:\delta(\mathcal{F}a, y)\to \varphi(\mathcal{GF}a, \mathcal{G}y)=\|X(Gy-GXa-Gb)\|=\|X(Gy-a)-XGb\|$,
and $\omega_a = \varphi(a, \mathcal{GF}a)=\|X\{G(Xa+b)-a\}\|=\|Xa+XGb-Xa\|=\|XGb\|$.
From the triangle inequality $\|X(Gy-a)-XGb\|+\|XGb\| \ge \|X(Gy-a)\|$, we obtain that the inequality (\ref{adj-02a1}) holds.
Moreover, if $b=b_{\perp}$, then $XGb=0$,
and hence equation (\ref{adj-02b1}) follows.
In this case, the $\Lambda$-unit $\omega_a$ becomes the genuine unit $\omega_{a,b_{\perp}}=0$ of the adjunction in the sense of ${\bf V}$-natural transformation.
\hfill $\square$

\bigskip
The Gauss-Markov adjunction expressed in Proposition {\bf (GM-2)} is summarized by the following diagram:
\begin{equation}
\xymatrix{
I \otimes \delta(\mathcal{F}_{b_{\perp}} a, y) \ar@<0.5ex>[d] \ar[rr]^{\bar \omega_{a,b_{\perp}} \otimes \mathcal{G} \qquad}	&& \varphi(a, \mathcal{GF}_{b_{\perp}} a) \otimes \varphi(\mathcal{GF}_{b_{\perp}} a, \mathcal{G} y )\ar[d]^{\cong}\\
\delta(\mathcal{F}_{b_{\perp}}a, y)	\ar@<0.5ex>[u]^{\cong} \ar@<0.5ex>[d]							&&\varphi(a, \mathcal{G} y)				\\
\delta(\mathcal{F}_{b_{\perp}}a, y) \otimes I \ar@<0.5ex>[u]^{\cong} \ar[rr]_{\mathcal{G} \otimes \bar \omega_{a,b_{\perp}} \qquad}	&& \varphi(\mathcal{GF}_{b_{\perp}} a, \mathcal{G} y ) \otimes \varphi(a, \mathcal{GF}_{b_{\perp}} a) \ar[u]_{\cong} \; .
}
\label{adj-02a3}
\end{equation}

\paragraph{Proposition (GM-3)}
There exists a $\Lambda$-natural transformation $\bar \varepsilon: \mathcal{FG} \to \mathcal{I}$,
\begin{equation}
\bar \varepsilon_y : \Lambda \to {\rm Hom}_{\bf Data}(\mathcal{FG}y, y)=\delta(\mathcal{FG}y, y)=\varepsilon_y,
\label{adj-03-unit}
\end{equation}
which we refer to as a \emph{$\Lambda$-counit}, with the following property.
For any objects $a \in {\rm Ob}({\bf Prm})$ and $y \in {\rm Ob}({\bf Data})$, and for a Hom-object $\delta(\mathcal{F}a,y)$, 
there exists a Hom-object $\varphi(a,\mathcal{G}y)$ such that the following diagram commutes:
\begin{equation}
\xymatrix{
\Lambda \otimes \varphi(a, \mathcal{G} y) \ar@<0.5ex>[dd] \ar[rr]^{\bar \varepsilon_y \otimes \mathcal{F} \qquad}	&& \delta(\mathcal{FG}y, y) \otimes \delta(\mathcal{F} a, \mathcal{FG} y)\ar[d]^{composition}\\
																		&&\delta(\mathcal{F} a, y)				\\
\varphi(a, \mathcal{G} y) \otimes \Lambda \ar@<0.5ex>[uu]^{\cong} \ar[rr]_{\mathcal{F} \otimes \bar \varepsilon_y \qquad}	&& \delta(\mathcal{F}a, \mathcal{FG} y) \otimes \delta(\mathcal{FG}y, y) \ar[u]_{composition} \; ,
}
\label{adj-03a2}
\end{equation}
where the right vertical arrow implies
\begin{equation}
\varepsilon_y + \delta(\mathcal{F}a,\mathcal{FG}y) \;\ge\; \delta(\mathcal{F}a,y).
\label{adj-03a1}
\end{equation}

In particular, if we select $b = \,^{\forall}b_{\perp} \in {\rm ker}(X^{\top})$, 
then $\varepsilon_{y,b_{\perp}}=0$, $\Lambda_{b_{\perp}}=I=0$, 
the equality in (\ref{adj-03a1}) holds, and $\varphi(a,\mathcal{G}y)$ is uniquely determined for any $a$, $y$, and $\delta(\mathcal{F}a,y)$; that is,
\begin{equation}
\varepsilon_{y,b_{\perp}} + \delta(\mathcal{F}_{b_{\perp}}a,\mathcal{F}_{b_{\perp}}\mathcal{G}y) \;=\; \delta(\mathcal{F}_{b_{\perp}}a,y),
\label{adj-03b1}
\end{equation}
which implies the isomorphism,
\begin{equation}
{\rm Hom}_{\bf Data}(\mathcal{F}_{b_{\perp}}\mathcal{G}y, y)\,\otimes\,{\rm Hom}_{\bf Data}(\mathcal{F}_{b_{\perp}}a,\mathcal{F}_{b_{\perp}}\mathcal{G}y)
\;\xrightarrow{\;\cong\;}\;
{\rm Hom}_{\bf Data}(\mathcal{F}_{b_{\perp}}a,y).
\label{adj-03b2}
\end{equation}
This is the universal mapping property of the counit $\varepsilon_{y,b_{\perp}}$, in the sense of ${\bf V}$-natural transformation.
\hfill $\square$

\paragraph{Proof (GM-3)}
From the definitions of the functors and the Hom-objects, we obtain
$\delta(\mathcal{F}a,y)=\|XG(y-Xa-b)\|=\|X(Gy-a)-XGb\|$,
$\mathcal{F}:\varphi(a, \mathcal{G}y) \to \delta(\mathcal{F}a, \mathcal{FG}y)=\|XG\left[\{X(Gy)+b\}-(Xa+b)\right]\|=\|X(Gy-a)\|$,
and $\varepsilon_y = (\mathcal{FG}y, y)=\|XG(y-XGy-b)\|=\|-XGb\|$.
From the triangle inequality $\|-XGb\|+\|X(Gy-a)\| \ge \|X(Gy-a)-XGb\|$, we obtain that the inequality (\ref{adj-03a1}) holds.
Moreover, if $b=b_{\perp}$, then $XGb=0$,
and hence equation (\ref{adj-03b1}) follows.
In this case, the $\Lambda$-counit $\varepsilon_y$ becomes the genuine unit $\varepsilon_{y,b_{\perp}}=0$ of the adjunction in the sense of ${\bf V}$-natural transformation.
\hfill $\square$

\bigskip
The Gauss-Markov adjunction expressed in Proposition {\bf (GM-3)} is summarized by the following diagram:
\begin{equation}
\xymatrix{
I \otimes \varphi(a, \mathcal{G} y) \ar@<0.5ex>[d] \ar[rr]^{\bar \varepsilon_{y,b_{\perp}}  \otimes \mathcal{F}_{b_{\perp}} \qquad}	&& \delta(\mathcal{F}_{b_{\perp}}\mathcal{G}y, y) \otimes \delta(\mathcal{F}_{b_{\perp}} a, \mathcal{F}_{b_{\perp}}\mathcal{G} y)\ar[d]^{\cong}\\
\varphi(a, \mathcal{G} y) \ar@<0.5ex>[u]^{\cong} \ar@<0.5ex>[d]							&&\delta(\mathcal{F}_{b_{\perp}} a, y)				\\
\varphi(a, \mathcal{G} y) \otimes I \ar@<0.5ex>[u]^{\cong} \ar[rr]_{\mathcal{F}_{b_{\perp}} \otimes \bar \varepsilon_{y,b_{\perp}}  \qquad}	&& \delta(\mathcal{F}_{b_{\perp}}a, \mathcal{F}_{b_{\perp}}\mathcal{G} y) \otimes \delta(\mathcal{F}_{b_{\perp}}\mathcal{G}y, y) \ar[u]_{\cong} \; .
}
\label{diag-01b}
\end{equation}

\section{Gradient Descent and Categorical Limits}
The Gauss-Markov adjunction provides, as it were, a comprehensive ``class-level'' understanding of how various possible training data vectors $y$ correspond to parameter values $a$, 
grounded in the structural formulation of regression.  
In contrast, assigning the optimized parameter vector $a^*$ and residual vector $r^*$ to a particular target vector $y$ captures an ``instance-level'' aspect of regression, representing the actual computational outcome.

In this section, we reinterpret gradient descent from a categorical perspective, 
showing how the Gauss-Markov adjunction clarifies the structural nature of convergence in iterative optimization.
Specifically, we reformulate the update sequences of parameters and residuals as diagrams within suitable Lawvere-enriched subcategories, 
and demonstrate that their categorical limits correspond to the optimal solutions $a^*$ and $r^*$.

\subsection{Optimality via Orthogonal Decomposition}
The regression problem admits a natural orthogonal decomposition of the target vector $y \in \mathbb{R}^n$:
$y = XGy + (I - XG)y = Xa^* + r^*$,
where $G$ is the Moore-Penrose inverse of $X$ (see {\bf Appendix C}).  
This decomposition uniquely determines the parameter vector and residual vector as
\begin{equation}
a^* := Gy \quad \text{and} \quad r^* := (I - XG)y.
\label{opt-01}
\end{equation}
Depending on the rank of $X$, the pair $(a^*,r^*)$ realizes optimality in complementary senses.  
If ${\rm rank}(X)=m < n$ (i.e., $X$ has full column rank), then $r^*$ is the residual of minimum norm and $a^*$ is the corresponding OLS estimator.  
If ${\rm rank}(X)=n \le m$ (i.e., $X$ has full row rank), then $r^*=0$ and $a^*$ is the minimum-norm solution among all parameter vectors fitting the data.  
In the general rank-deficient case ($\mathrm{rank}(X)<\min(m,n)$), the Moore-Penrose inverse consistently yields (\ref{opt-01}), which provide the generalized optimal solution.

\subsection{Diagrams induced from Iterative Updates}
Consider the ordinary gradient descent procedure for multiple regression (see also {\bf Appendix B}).
The iterative updates for the parameter $a$ and the residual $r$, for $i \in \mathbb{N}$,
\begin{subequations} 
    \begin{align}
	&a_{i+1} = a_i + \eta X^{\top}(y-Xa_i)=:f_a(a_i) \label{update-01a}\\
	&r_{i+1} = r_i - \eta XX^{\top}r_i=:f_r(r_i) \label{update-01b}
    \end{align}
\end{subequations}
are induced from
$a_{i+1} := a_i - \eta \nabla_{a_i} L(a_i)$, $L(a_i):=\|y-Xa_i\|$.
The iterative updates of gradient descent converge when the step size coefficient $\eta$ satisfies $0 < \eta < 2/\beta_{\max}(XX^{\top})$.
These recurrences (\ref{update-01a}) and (\ref{update-01b}) are related by 
\begin{equation}
r_i=y-Xa_i.
\label{r_i-01}
\end{equation}
Starting from a pair of initial states $(a_1,r_1)$, these recurrences generate $\{a_1,a_2,\dots,a_{\infty}\}$ and $\{r_1,r_2,\dots,r_{\infty}\}$,
where the analytic limits $a_{\infty}=\lim\limits_{i\rightarrow \infty} a_i$ and $r_{\infty}=\lim\limits_{i\rightarrow \infty} r_i$ appear as the unique fixed points of the recurrences, such that
\begin{equation}
a_{\infty}=f_a(a_{\infty}) \quad \text{and} \quad r_{\infty}=f_r(r_{\infty}).
\label{fixed-01}
\end{equation}
Moreover, we define the sequences of $S_{\delta}$ and $S_{\varphi}$ as
\begin{subequations} 
    \begin{align}
			&S_{\delta}:=\{r_i \mid r_{i+1}=f_r(r_i),i = 1,\dots,{\infty} \} \label{seq-01a}\\
	\text{and}\quad &S_{\varphi}:=\{G(y-Xa_i) \mid a_{i+1}=f_a(a_i),i = 1,\dots,{\infty} \}. \label{seq-01b}
    \end{align}
\end{subequations}
In categorical sense, for all $i \in \mathbb{N}$, $r_i \in {\rm Ob}({\bf Data})$ and $G(y-Xa_i) \in {\rm Ob}({\bf Prm})$,
therefore the sequences $S_{\varphi}$ and $S_{\delta}$ induce the diagrams 
$\mathcal{D}_{\delta}:{\bf J}\to {\bf Data}$ and $\mathcal{D}_{\varphi}=\mathcal{G}\circ\mathcal{D}_{\delta}:{\bf J}\to {\bf Prm}$, where ${\bf J}$ is the index category.
We refer to these as the diagrams of the gradient descent.
Let ${\bf S_{\varphi}}$ and ${\bf S_{\delta}}$ denote the subcategories of ${\bf Prm}$ and ${\bf Data}$ generated by the images of $\mathcal{D}_{\varphi}$ and $\mathcal{D}_{\delta}$.

\subsection{Cones of Diagrams of Gradient Descent}
Let us consider the cones of the diagrams of gradient descent, 
$\mathcal{D}_{\delta}$ and $\mathcal{D}_{\varphi}$, denoted by ${\rm Cone}(\mathcal{D}_{\delta})$ and ${\rm Cone}(\mathcal{D}_{\varphi})$.
In general, a cone ${\rm Cone}(\mathcal{D})$ of a diagram $\mathcal{D}:{\bf J}\to{\bf C}$ with vertex $c\in {\rm Ob}({\bf C})$ 
consists of a family of morphisms $\{\nu_i:c\to \mathcal{D}(i)\}_{i\in {\rm Ob}({\bf J})}$ 
such that for every morphism $+j:i\to i+j$ in ${\bf J}$ the condition 
$\mathcal{D}(+j)\circ \nu_i = \nu_{i+j}$ holds.

In our case, 
${\rm Cone}(\mathcal{D}_{\delta})$ (resp. ${\rm Cone}(\mathcal{D}_{\varphi})$ ) consists of a vertex $y_c\in {\rm Ob}({\bf Data})$ (resp. $Gy_c\in {\rm Ob}({\bf Prm})$) 
together with a family of Hom-objects $\{\varphi(y_c, r_i) \}_{i\in{\bf J}}$ with $r_i\in {\rm Ob}({\bf S_{\delta}})$ (resp.  $\{\varphi(Gy_c, G(y-Xa_i)) \}_{i\in{\bf J}}$ with $G(y-Xa_i)\in {\rm Ob}({\bf S_{\varphi}})$),
such that for every $i\le j$ in ${\bf J}$ the enriched cone condition holds:
\begin{eqnarray}
	\delta(r_i,r_j)\;+\;\delta(y_c,r_i) &\ge& \delta(y_c,r_j), \label{cone-01a}\\
	\text{resp.}\quad \varphi(G(y-Xa_i),G(y-Xa_j))&+&\varphi(a_c,G(y-Xa_i)) \nonumber \\
	&\ge& \varphi(a_c,G(y-Xa_j)). \label{cone-01b}
\end{eqnarray}
Since the definitions of the Hom-objects of ${\bf Data}$ and  ${\bf Prm}$ satisfy the triangle inequality, 
the conditions (\ref{cone-01a}) and (\ref{cone-01b}) are automatically fulfilled 
for every $y_c \in {\rm Ob}({\bf Data})$ and every $Gy_c \in {\rm Ob}({\bf Prm})$. 
Consequently, any object in ${\bf Data}$ or ${\bf Prm}$ can serve as a vertex of a cone of the corresponding diagram of gradient descent.

\subsection{Categorical Limits of Diagrams of Gradient Descent}
In the previous subsection we introduced the cones of the diagrams of gradient descent, 
$\mathcal{D}_{\delta}:{\bf J}\to{\bf Data}$ and $\mathcal{D}_{\varphi}:{\bf J}\to{\bf Prm}$. 
By definition, any object in ${\bf Data}$ or ${\bf Prm}$ can serve as a vertex of such a cone, 
since the enriched triangle inequalities are always satisfied. 
However, among these cones there exist universal ones, which characterize the categorical limits of the diagrams, 
$\lim\limits_{\longleftarrow}\mathcal{D}_{\delta}$ and $\lim\limits_{\longleftarrow}\mathcal{D}_{\varphi}$.
In general, a limit $\lim\limits_{\longleftarrow}\mathcal{D}$ of a diagram $\mathcal{D}:{\bf J}\to{\bf C}$ is a cone $(z,\{\zeta_i:z\to\mathcal{D}(i)\})$ 
that is universal in the sense that for any other cone $(c,\{\nu_i:c\to \mathcal{D}(i)\})$ 
there exists a unique morphism $u:c\to z$ such that $\zeta_i\circ u = \nu_i$ for all $i\in{\rm Ob}({\bf J})$.
This definition emphasizes that the categorical limit is determined not merely by convergence in the analytic sense, 
but by a universal mapping property in the categorical sense.

\paragraph{Limit of residuals}
We show that, for the residual diagram $\mathcal{D}_{\delta}$, 
the minimum-norm residual $r^* = (I - XG)y$ in (\ref{opt-01}) serves as the vertex of the universal cone, that is, 
the limit object $r^* = \lim\limits_{\longleftarrow}\mathcal{D}_{\delta}$, 
and moreover that the fixed point $r_{\infty} = \lim\limits_{i\rightarrow \infty} r_i$ obtained as the analytic limit coincides with $r^*$.

First, any $y_c \in {\rm Ob}({\bf Data})$ gives rise to a cone $(y_c,\{\delta(y_c,r_i)\}_{i\in{\bf J}})$ of $\mathcal{D}_{\delta}$,
and in particular $(r^*,\{\delta(r^*,r_i)\}_{i\in{\bf J}})$ is also a cone of $\mathcal{D}_{\delta}$. 
Secondly, for all $i \in {\bf J}$ and for any cone $(y_c,\{\delta(y_c,r_i)\}_{i\in{\bf J}})\in{\rm Cone}(\mathcal{D}_{\delta})$, 
the Hom-object $\delta(y_c, r^*)$ is well-defined, and the cone 
$(r^*,\{\delta(r^*,r_i)\}_{i\in{\bf J}})$ satisfies the triangle inequality
\begin{equation}
\delta(r^*, r_i) + \delta(y_c, r^*) \;\ge\; \delta(y_c, r_i),
\label{cone-r-01}
\end{equation}
since these distances are the Hom-objects in ${\bf Data}$.
Finally, 
from $Gr^*=0 \Leftrightarrow r^*\in {\rm ker}(G)$, $\delta(y_c, r^*)=\|XG(r^*-y_c)\|=\|-XGy_c\|$,
and from (\ref{update-01b}),
$r_{\infty} = f_r(r_{\infty})\Leftrightarrow X^{\top}r_{\infty}=0 \Leftrightarrow r_{\infty} \in {\rm ker}(X^{\top})={\rm ker}(G)$,
and $r_{\infty}-r^* \in {\rm ker}(G)$, thus $\delta(r^*, r_{\infty})=\|XG(r_{\infty}-r^*)\|=0$, and $\delta(y_c, r_{\infty})=\|XG(r_{\infty}-y_c)\|=\|-XGy_c\|$. 
Therefore, from (\ref{cone-r-01}),
\begin{eqnarray}
\delta(y_c, r^*) 
&=&\|-XGy_c\|\nonumber\\
&\;\ge\;& \delta(y_c, r_i) - \delta(r^*, r_i) \nonumber\\
&\xrightarrow[i\to \infty]{}& \delta(y_c, r_{\infty}) - \delta(r^*, r_{\infty})\nonumber\\
&=& \delta(y_c, r_{\infty})=\|-XGy_c\|,
\label{cone-r-02}
\end{eqnarray}
that is, the Hom-object $\delta(y_c, r^*)$ is uniquely determined and equals $\delta(y_c, r_{\infty})$ for each vertex $y_c$.
Moreover, substituting $r^*$ to (\ref{seq-01b}), we obtain $r^* = f_r(r^*)$ since $X^{\top}r^*=0$, thus $r^*=r_{\infty}$. 
Consequently, the categorical limit, the optimal residual, and the analytic limit all coincide:
\begin{equation}
\lim\limits_{\longleftarrow}\mathcal{D}_{\delta} \;=\; r^* \;=\; \lim\limits_{i\to\infty} r_i.
\label{3lim-r-01}
\end{equation}

\paragraph{Right adjoint preserves limits}
The Gauss-Markov functor $\mathcal{G}:{\bf Data}\to{\bf Prm}$ preserves limits, 
since right adjoints in general preserve limits (RAPL) \cite{Awodey2010}, 
and $\mathcal{G}$ is the right adjoint in the Gauss-Markov adjunction.
Indeed, applying the functor $\mathcal{G}$ to ${\rm Hom}_{\bf Data}(y_c, r_{\infty})$ 
yields the following chain of Hom-objects:
\begin{eqnarray}
{\rm Hom}_{\bf Prm}(\mathcal{G}y_c, \mathcal{G}r_{\infty}) 
&=&{\rm Hom}_{\bf Prm}(\mathcal{G}y_c, \mathcal{G}(\lim\limits_{\longleftarrow} \mathcal{D}_{\delta})) \nonumber \\
&=& {\rm Hom}_{\bf Data}(\mathcal{F}_{b_{\perp}}\mathcal{G}y_c ,\lim\limits_{\longleftarrow} \mathcal{D}_{\delta})\nonumber \\
&=& \lim\limits_{\longleftarrow}{\rm Hom}_{\bf Data}(\mathcal{F}_{b_{\perp}}\mathcal{G}y_c , r_i)\nonumber \\
&=& \lim\limits_{\longleftarrow}{\rm Hom}_{\bf Prm}(\mathcal{G}y_c, \mathcal{G}r_i) \nonumber \\
&=& {\rm Hom}_{\bf Prm}(\mathcal{G}y_c, \lim\limits_{\longleftarrow}\mathcal{GD}_{\delta})  \nonumber \\
&=& {\rm Hom}_{\bf Prm}(\mathcal{G}y_c, \lim\limits_{\longleftarrow}\mathcal{D}_{\varphi}) .
\end{eqnarray}
Moreover, $r_{\infty}=r^*=(I-XG)y$ ($\because$ (\ref{opt-01})), thus ${\rm Hom}_{\bf Data}((I-XG)y, r_{\infty})=0$.
Since $\mathcal{G}:0\to G0 = 0$,
\begin{eqnarray}
{\rm Hom}_{\bf Prm}(\mathcal{G}(I-XG)y, \lim\limits_{\longleftarrow}\mathcal{D}_{\varphi}) 
&=& \varphi(\mathcal{G}(I-XG)y, \mathcal{G}r_{\infty}) \nonumber \\
&=& \|X(Gr_{\infty}-G(I-XG)y)\| \nonumber \\
&=& \|X(Gy-GXa_{\infty})\| \nonumber \\
&=& \|X(Gy-a_{\infty})\|=0
\end{eqnarray}
Consequently, we obtain
\begin{equation}
\lim\limits_{\longleftarrow}\mathcal{D}_{\varphi} = Gy-a_{\infty} = 0,
\end{equation}
and
\begin{equation}
a^*=Gy=a_{\infty}.
\end{equation}

\section{Discussion}
In this paper, we have presented the Gauss-Markov Adjunction (GMA), a categorical reformulation of multiple linear regression, which is the most fundamental form of supervised learning. 
This framework explicitly describes the dual flow of information between residuals and parameters, and demonstrates that their relationship is governed by the adjunction structure.

Our proposed GMA framework can be understood as an extension of denotational semantics, originally developed for programs with formal syntax, to the domain of supervised learning systems. 
Denotational semantics is a methodology that clarifies the meaning of a program by assigning a mathematical semantic object to each syntactic construct \cite{Slonneger-Kurtz1994}. 
In our formulation, we associate the core components of multiple regression---data, parameters, and residuals---with categorical objects and morphisms, and explicitly characterize their structure as an adjunction. 
This categorical approach offers a pathway toward constructive semantic understanding in the context of statistical machine learning, including modern AI systems.

A representative success of denotational semantics grounded in category theory is the correspondence between typed $\lambda$-calculi and Cartesian Closed Categories (CCCs) \cite{Lambek-Scott1988}. 
A CCC is a category that has all finite products and exponentials, and it is characterized by the adjunction $(-) \times A \dashv (-)^A$, 
through which the semantics of $\lambda$-calculus is constructively interpreted. 
GMA represents a novel application within this tradition, aiming to endow statistical machine learning models with a similarly structured semantic interpretation.

Categorical modeling, understood as a form of semantic structuring, can be positioned as a framework that contributes to the realization of Explicability as a core principle of AI. 
Explicability does not merely demand access to source code, but requires that the behavior and structure of AI systems be made understandable and explainable at an appropriate level of abstraction. 
At its core, this principle encompasses both intelligibility and interpretability as essential components.

In modern AI models such as deep learning, decisions about when training is complete and how to evaluate the validity of outputs often rely on empirical judgment or operational heuristics. 
As a result, it is not straightforward to provide a semantic framework for these models. 
In particular, no principled framework has yet been established that connects the convergence of parameters with the semantic validity of outputs.

To be sure, our framework does not claim to guarantee optimality or convergence across all aspects of the learning process in machine learning. 
Instabilities resulting from the non-convexity of objective functions or the complexity of model architectures may still fall outside the scope of what the GMA framework can directly assure.

Nevertheless, as this study demonstrates, 
introducing the adjunction as a semantic structure between the components of a supervised learning model enables a structural understanding that transcends specific numerical behaviors. 
This reveals a new significance in applying an extended form of denotational semantics to machine learning models.

\section{Conclusion}
In this paper, we have proposed a semantically grounded reformulation of the multiple linear regression model, arguably the most fundamental form of supervised learning, based on a categorical construction. 
The central result of this work is the explicit identification of an adjunction that arises between data and parameters, mediated by the residuals. 
We refer to this structure as the Gauss-Markov Adjunction (GMA). 
This framework enables a clear and compositional categorical understanding of the information flow involved in the learning process of regression models.

The theoretical contribution of this study lies in offering a constructive, category-theoretic semantic interpretation for supervised learning systems. 
In particular, the GMA clarifies the structural relationships among data, parameters, and residuals by mapping them explicitly onto categorical objects and morphisms. 
In this sense, the approach presented in this paper can be regarded as an extended application of denotational semantics.

Traditionally, denotational semantics has been developed as a formal method for the interpretation of programming languages, 
and has had considerable influence on functional programming and type theory. 
Our approach represents a novel extension of this tradition, providing a theoretical foundation for enhancing the intelligibility and interpretability of AI systems, 
particularly in the context of learning architectures.

Furthermore, the framework presented in this study extends beyond multiple regression and has the potential to be applied to more general supervised learning systems. 
Future work will explore categorical extensions to neural networks and other models, with the aim of developing a more comprehensive semantic framework for AI.

\section*{Appendix}
\paragraph{A. Regression}
Under the condition ${\rm rank}(X) = m \le n$ (i.e., $X$ has full column rank), 
the regression minimizes the $L^2$-norm of the residual vector $r$.
The optimal solution $a^* = Gy$ is the ordinary least squares (OLS) estimator, 
where $G = G_L= (X^{\top}X)^{-1}X^{\top}$ is the left inverse of $X$ 
(i.e., $G_LX = I$, while $XG_L \neq I$, which represents the projection matrix).
The best-fit model for the given dataset $(X, y)$ is $y = Xa^* + r_{\perp}$,
where $r_{\perp} = y - Xa^* = (I - XG)y$ is the minimum residual.
Under the condition ${\rm rank}(X) = n \le m$ (i.e., $X$ has full row rank), 
the solution retains the form $a^* = Gy$, while the residual vanishes ($r=0$). 
When $n = m$, $X$ is a nonsingular matrix with $G = X^{-1}$. 
When $n < m$, the given data alone are insufficient to uniquely determine the parameters. 
In this case, a solution is obtained by imposing the additional condition that the norm of the parameter vector $\|a\|$ is minimized. 
The corresponding expression for $G$ is $G = G_R= X^{\top}(XX^{\top})^{-1}$, which satisfies $XG_R = I$ and $G_RX \neq I$.
Under the condition ${\rm rank}(X) < \min(m,n)$ (i.e., $X$ is rank-deficient), 
the solution also retains the form $a^* = Gy$, 
where $G = X^+$ denotes the Moore-Penrose inverse of $X$. 
In this case, $XG \neq I$ and $GX \neq I$, that is, $G$ is neither a left nor a right inverse.
In all of the above cases, regardless of the relative sizes of $m$ and $n$, $G$ is the Moore-Penrose inverse of $X$, 
which is uniquely determined as a generalized inverse of $X$.
Harville's outstanding textbook \cite{Harville2008} offers a comprehensive treatment of matrix algebra for statistics, 
which serves as a valuable reference for understanding these properties in detail.

\paragraph{B. Iterative Method}
Regression can be interpreted as a supervised learning system that learns the optimal parameter $a^*$ which enables accurate predictions for the training data $X$, 
with the value of $y$ serving as the supervisory signal.
The objective function is often expressed as $L(a) = \frac{1}{2} \|r\|^2 = \frac{1}{2} \|y - Xa\|^2$.
Starting from an initial state $a_1$ (arbitrary when ${\rm rank}(X)=m \le n$ and $a_1=0$ when ${\rm rank}(X) \le n < m$), 
one can asymptotically approach $a^*$ using the iterative method $a_{i+1} = a_i - \eta \nabla_{a_i} L(a_i) \; (0 < \eta < \eta_{\max} \; ; \; i = 1,2,\cdots)$, 
and eventually reach $a^* = \lim\limits_{i\to \infty} a_i$.
Note that $\eta_{\max}:=2/\beta_{\max}(XX^{\top})$,
derived from $\|I-\eta XX^{\top}\|<1$,
where $\beta_{\max}(XX^{\top})=\|X\|^2$ denotes the maximum eigenvalue of $XX^{\top}$.

\paragraph{C. Generalized inverse and Moore-Penrose inverse}
For any $n\times m$ matrix $A$, any $m\times n$ matrix $G$ satisfying the condition (MP1) $AGA=A$ is referred to as a generalized inverse of $A$.
In general, there are infinitely many such matrices $G$. A matrix $G$ that satisfies, 
in addition to the condition (MP1), the conditions (MP2)-(MP4) is referred to as the Moore-Penrose inverse of $A$: 
(MP2) $GAG=A$, (MP3) $(AG)^{\top}=AG$, (MP4) $(GA)^{\top}=GA$. This inverse is uniquely determined for each matrix $A$. 
The notions of the generalized inverse and the Moore-Penrose inverse are defined for arbitrary matrices $A$, including those that are rank-deficient (${\rm rank}(A)<\min(n,m)$).

\paragraph{D. Left and right inverse}
When the matrix $A$ is full column rank, its generalized inverse $G$ becomes a left inverse ($GA=I$). 
When $A$ is row full rank, $G$ becomes a right inverse ($AG=I$). 
These cases are specified by the condition ${\rm rank}(A)=\min(n,m)$, and there exist infinitely many such matrices $G$. 
If condition (MP3) is imposed on the left inverse, $G$ becomes the Moore-Penrose inverse
($\because$ (MP1)$\wedge GA=I \Rightarrow GAGAG=GAG=G \Leftrightarrow$(MP2) and (MP3)$\wedge GA=I \Rightarrow (GA)^{\top}=I=GA \Leftrightarrow$(MP4)), 
which is uniquely determined and coincides with the solution obtained by the least squares method in regression. 
Similarly, if condition (MP4) is imposed on the right inverse, $G$ becomes the Moore-Penrose inverse, 
which is uniquely determined and coincides with the solution obtained under the minimum-norm constraint in regression. 
In the case where $A$ is nonsingular (${\rm rank}(A)=n=m$), we simply have $G = A^{-1}$.

\section*{Acknowledgements}
I am deeply grateful to the three anonymous reviewers for their contributions, which amount to co-authorship in spirit. 
One reviewer clearly explained why our previous definitions of the two categories were inadequate. 
Another suggested a more constructive possibility of improving these definitions by employing Lawvere-enriched categories. 
The third suggested that the conditions imposed on the training data matrix were unnecessary, 
and that the Gauss-Markov adjunction could more generally be defined through the Moore-Penrose inverse.

This work was supported by JSPS KAKENHI Grant Number JP25K03532.


\begin{thebibliography}{99}
\bibitem{Abbott+2024}
Abbott, V., Xu, T., Maruyama, Y. (2024). ``Category Theory for Artificial General Intelligence,'' 
In: {\it Artificial General Intelligence AGI 2024 Lecture Notes in Computer Science}, 14951. 
Springer. \url{https://doi.org/10.1007/978-3-031-65572-2_13}

\bibitem{Asilomar2017}
Asilomar AI Principles (2017). \url{https://futureoflife.org/open-letter/ai-principles/}

\bibitem{Awodey2010}
Awodey, S. (2008; 2010(2nd)) {\it Category Theory}, Oxford University Press.

\bibitem{Beauchamp-Childress1979} 
Beauchamp, T.L. and Childress, J.F. (1979). {\it Principles of Biomedical Ethics}. Oxford University Press.

\bibitem{Coeckelbergh2020}
Coeckelbergh, M. (2020). {\it AI Ethics}, MIT Press.

\bibitem{Cruttwell2021}
Cruttwell, G.S.H., Gavranovi\'{c}, B., Ghani, N, Wilson, P., and Zanasi, F. (2021).
``Categorical Foundations of Gradient-Based Learning,'' arXiv:2103.01931
\url{https://doi.org/10.48550/arXiv.2103.01931}

\bibitem{Harville2008}
Harville, D. A. (1997; 2008(2nd)) {\it Matrix Algebra from a Statistician's Perspective}. Springer.

\bibitem{He+2016}
He, K., Zhang, X., Ren, S., and Sun, J. (2016). 
``Deep Residual Learning for Image Recognition,''
{\it Proceedings of the IEEE Conference on Computer Vision and Pattern Recognition (CVPR)}, 770-778.
\url{https://doi.org/10.1109/CVPR.2016.90}

\bibitem{Heunen-Vicary2019}
Heunen, C. and Vicary, J. (2019). {\it Categories for Quantum Theory: An Introduction}. Oxford University Press.

\bibitem{Floridi2008}
Floridi, L. (2008). ``The method of levels of abstraction,'' {\it Minds and Machines}, 18(3)303-329. \url{https://psycnet.apa.org/doi/10.1007/s11023-008-9113-7}

\bibitem {Floridi2019}
Floridi, L. (2019). ``What the Near Future of Artificial Intelligence Could Be,'' Philosophy \& Technology, 32, 1-15. \url{https://doi.org/10.1007/s13347-019-00345-y}

\bibitem{Floridi+2018}
Floridi, L., Cowls, J., Beltrametti, M., Chatila, R., Chazerand, P., Dignum, V., Luetge, C., Madelin, R., Pagallo, U., Rossi, F., Schafer, B., Valcke, P., and Vayena, E. (2018).
``AI4People-An Ethical Framework for a Good AI Society: Opportunities, Risks, Principles, and Recommendations,'' 
Minds and Machines, 28, 689-707. \url{https://doi.org/10.1007/s11023-018-9482-5}

\bibitem {Floridi-Cowls2019}
Floridi, L. and Cowls, J. (2019). ``A Unified Framework of Five Principles for AI in Society,'' {\it Harvard Data Science Review}, 1(1). \url{https://doi.org/10.1162/99608f92.8cd550d1}

\bibitem{Fong2019}
Fong, B., Spivak, D.I., and Tuy\'{e}ras, R. (2019)
``Backprop as Functor: A compositional perspective on supervised learning,'' arXiv:1711.10455
\url{https://doi.org/10.48550/arXiv.1711.10455}

\bibitem {Gavranovic2024}
Bruno Gavranovi\'{c}, Paul Lessard, Andrew Dudzik, Tamara von Glehn, Jo\~{a}o G. M. Araujo, Petar Veli\v{c}kovi\'{c} (2024).
``Position: Categorical Deep Learning is an Algebraic Theory of All Architectures,'' arXiv:2402.15332
\url{https://doi.org/10.48550/arXiv.2402.15332}

\bibitem{IEEE2017}
IEEE Initiative on Ethics of Autonomous and Intelligent Systems:
Ethically Aligned Design: A Vision for Prioritizing Human Well-being with Autonomous and Intelligent Systems, Version 2. (2017). 
\url{https://standards.ieee.org/wp-content/uploads/import/documents/other/ead\_v2.pdf}

\bibitem{Kamiura2025}
Kamiura, M. (2025). ``The Four Fundamental Components for Intelligibility and Interpretability in AI Ethics,'' The American Philosophical Quarterly,
62(2)103-112. \url{https://doi.org/10.5406/21521123.62.2.01}

\bibitem{Kelly1982}
Kelly, G. M. (1982) {\it Basic concepts of enriched category theory}, London Mathematical Society Lecture Note Series, 64. Cambridge University Press.

\bibitem{Lambek-Scott1988}
Lambek, J. and Scott, P.J. (1988) {\it Introduction to Higher Order Categorical Logic}, Cambridge University Press.

\bibitem{Lawvere1973}
Lawvere, F. W. (1973) ``Metric spaces, generalized logic, and closed categories,'' 43, 135-166.
\url{https://doi.org/10.1007/BF02924844}

\bibitem{MacLane1978}
Mac Lane, S. (1978; 1998(2nd)). {\it Categories for the Working Mathematician}. Springer.

\bibitem{Pan2024}
Pan, W. (2024). ``Token Space: A Category Theory Framework for AI Computations,'' arXiv:2404.11624
\url{https://doi.org/10.48550/arXiv.2404.11624}

\bibitem{Sheshmani-You2021}
Sheshmani, A. and You, Y-Z. (2021). ``Categorical representation learning: morphism is all you need,'' {\it Machine Learning: Science and Technology}, 3, 015016.
\url{https://iopscience.iop.org/article/10.1088/2632-2153/ac2c5d}

\bibitem{Shiebler+2021}
Shiebler, D., Gavranovi\'{c}, B., and Paul Wilson, P. (2021). ``Category Theory in Machine Learning,'' arXiv:2106.07032
\url{https://doi.org/10.48550/arXiv.2106.07032}

\bibitem{Slonneger-Kurtz1994}
Slonneger, K. and Kurtz, B.L. (1994) {\it Formal Syntax and Semantics of Programming Languages : A Laboratory Based Approach}, Addison-Wesley Publishing.

\bibitem{Steingartner2025}
Steingartner, W. (2025). ``Perspectives of semantic modeling in categories,'' Journal of King Saud University Computer and Information Sciences, 37(19). 
\url{https://doi.org/10.1007/s44443-025-00010-9}

\bibitem{Ursin+2023}
Ursin, F., Lindner, F., Ropinski, T., Salloch, S. and Timmermann, C. (2023). 
``Levels of explicability for medical artificial intelligence: What do we normatively need and what can we technically reach?,'' {\it Ethik in der Medizin}, 35, 173-199.
\url{https://doi.org/10.1007/s00481-023-00761-x}

\bibitem{Vaswani+2017}
Vaswani, A., Shazeer, N., Parmar, N., Uszkoreit, J., Jones, L., Gomez, A.N., Kaiser, L., and Polosukhin, I. (2017).
 ``Attention is all you need,'' {\it Proceedings of the 31st International Conference on Neural Information Processing Systems 2017}, 6000-6010.
\url{https://dl.acm.org/doi/10.5555/3295222.3295349}

\bibitem{Xu+2022}
Xu, T. and Maruyama, Y. (2022). ``Neural String Diagrams: A Universal Modelling Language for Categorical Deep Learning,''
In: {\it Artificial General Intelligence AGI 2021 Lecture Notes in Computer Science}, 13154. Springer. 
\url{https://doi.org/10.1007/978-3-030-93758-4_32}

\end{thebibliography}
\end{document}